\pdfoutput=1

\documentclass[11pt]{article}

\usepackage[preprint]{acl}

\usepackage{times}
\usepackage{latexsym}

\usepackage[T1]{fontenc}

\usepackage[utf8]{inputenc}

\usepackage{microtype}

\usepackage{inconsolata}

\usepackage{graphicx}

\usepackage{longtable}
\usepackage{amsfonts,amssymb} 
\usepackage{booktabs}
\usepackage{multirow}
\usepackage{color}
\usepackage{cite} 
\usepackage{hyperref}
\title{PGA-SciRE: Harnessing LLM on Data Augmentation for Enhancing Scientific Relation Extraction}

\author{Yang Zhou$^1$ , Shimin Shan$^{1}$\thanks{Co-corresponding Author.} , Hongkui Wei$^2$ ,  Zhehuan Zhao$^{1*}$ , Wenshuo Feng$^1$ \\
  $^1$Software school, Dalian University of Technology, Dalian\ 116620 \\
  $^2$State Key Laboratory of Intelligent Manufacturing System Technology, \\
  Beijing Institute of Electronic System Engineering, Beijing\ 100854 \\
  \texttt{\{zy22217034,fengws\}@mail.dlut.edu.cn} \\
  \texttt{weihongkui89@hotmail.com} \\
  \texttt{ \{ssm,z.zhao\}@dlut.edu.cn}
  }

\begin{document}
\maketitle
\begin{abstract}
    Relation Extraction (RE) aims at recognizing the relation between pairs of entities mentioned in a text. Advances in LLMs have had a tremendous impact on NLP. In this work, we propose a textual data augmentation framework called PGA for improving the performance of models for RE in the scientific domain. The framework introduces two ways of data augmentation, utilizing a LLM to obtain pseudo-samples with the same sentence meaning but with different representations and forms by paraphrasing the original training set samples. As well as instructing LLM to generate sentences that implicitly contain information about the corresponding labels based on the relation and entity of the original training set samples. These two kinds of pseudo-samples participate in the training of the RE model together with the original dataset, respectively. The PGA framework in the experiment improves the F1 scores of the three mainstream models for RE within the scientific domain. Also, using a LLM to obtain samples can effectively reduce the cost of manually labeling data.
\end{abstract}

\section{Introduction}

The leap forward in LLMs and novelty ICL methods has improved the performance and efficiency of NLP tasks, and has created new opportunities for solving them\citep{brown2020language}. Plenty of work has investigated LLMs’ performance, the capabilities of reasoning and interpretation, and presented systematic analysis\citep{li2023evaluating}. \citet{wei2023zeroshot} provide a comprehensive study and detailed analysis of the capabilities of LLMs by examining the performance on IE tasks. \citet{han2023information} propose a soft-matching evaluation strategy as criteria for IE on LLM to reflect the performance more accurately. All these works demonstrate the great potential of LLM.

Relation Extraction (RE) is the task of obtaining structured knowledge from unstructured text and identifying semantic relations between entities in text. Usual methods are Utilizing ICL and CoT approaches by considering RE as a conditional text generation task \citep{wan-etal-2023-gpt, ma-etal-2023-chain-thought} or a question and answer task \citep{zhang-etal-2023-aligning, li-etal-2023-revisiting-large}. Input test samples, examples, task descriptions, and answers are given directly by the LLM. These methods are mainly applied in low-shot settings, while the flexibility of the language in which the LLM outputs are given leads to the challenge of evaluating the results \citep{wadhwa-etal-2023-revisiting}. At the same time, these methods are maybe not good at working with RE in the scientific domain since specialized concepts are consistently defined and proposed every year, and the corpus on which the LLM is trained does not contain these new acronyms and concepts.

Acquiring large-scale and high-quality data with annotated information for various tasks has long been a very challenging and costly endeavor \citep{beltagy-etal-2019-scibert}. Especially in the domain of scientific knowledge graphs, since scientific datasets tend to face unique challenges. Scientific datasets may be more noisy; and contain a large number of specialized vocabularies, as well as acronyms specific to particular concepts, making it difficult for models to learn and generalize; and these datasets tend to lack extensive annotation due to the time and labor costs involved in organizing them.

Apart from model development, RE always involves labeled data. The performance of models not only in the few-shot setting, but also in the full-sample training setting is heavily influenced by the quality and quantity of the training data. While the cost of labeling data in the scientific domain is high, it is a very common problem in practice. LLM brings new approaches to solving the problem by making it possible to generate labeled data \citep{moller2024parrot}. In some work, LLM-based data augmentation methods mainly focus on solving text classification task \citep{dai2023auggpt, piedboeuf-langlais-2023-chatgpt, tang2023does}, while RE task contains richer predefined label information \citep{xu2023unleash}, as well as a wider classification space, and it is more challenging on datasets in the scientific domain. In addition, these works are dedicated to RE in the few-shot setting, where the method only needs to produce a small number of samples for the model to perform well. In contrast, the usual full-sample model requires data of a size and quality that far exceeds that of a few-shot setting.

To this end, we start from the data side by using the proprietary vocabulary of scientific domains that are difficult to understand as input information for LLMs, compensating for their knowledge in this domain and allowing them to synthesize additional samples of in-domain data. This work proposes the framework called PGA (Paraphrasing and Generating Augmentation) for enhancing RE in scientific domains, which utilizes GPT-3.5 to synthesize paraphrased and generated pseudo-samples to enhance the performance of the RE model, the contributions as follows:

\begin{itemize}
  \item Our research shows that LLM is able to generate superior, labeled pseudo-samples for RE in scientific domains without manual labeling through simple but effective prompt we devised.
  
  \item The pseudo-samples obtained through the PGA framework employing paraphrasing and generating method can lead to improvement of the F1 scores for several mainstream RE models, respectively.
\end{itemize}

\section{Related Work}

\textbf{Relation Extraction}\quad RE aims at identifying relations between pairs of entities mentioned in a text. The mainstream methods can be categorized into pipelined and joint methods. (1) Pipelined methods deal with two subtasks, NER and RE, sequentially, but suffer from error propagation \citep{ye-etal-2022-packed, zhong-chen-2021-frustratingly, miwa-bansal-2016-end}. (2) Joint approaches can alleviate this problem by processing two subtasks simultaneously. \citet{yan-etal-2021-partition, wang-etal-2021-unire, wang-lu-2020-two} process both tasks as tagging entries of a table. \citet{yan-etal-2023-joint} utilize HGNN for higher-order modeling and designed a joint ERE model based on the pipelined model of \citep{ye-etal-2022-packed}. In addition, \citet{ren-etal-2023-covariance} use a covariance optimization method to minimize the covariance of the features, which enhances the characterization of the features across both pipeline and joint models.
\\
\textbf{Data augmentation}\quad Paraphrasing-based data augmentation methods rewrite or rephrase the original text to obtain new text with the same meaning but different forms of expression, and paraphrasing techniques are widely used in NLP tasks \citep{kumar-etal-2019-submodular, okur2022data}, including RE \citep{yu-etal-2020-improving}. Due to the rise of LLM, paraphrasing techniques also flourish. \citet{piedboeuf-langlais-2023-chatgpt} investigate the performance of ChatGPT for paraphrasing data on a text categorization task.

Generating-based data augmentation requires that the generated data should be as close as possible to the data distribution of the original dataset and maintain similar information as the original data. \citet{meng2022generating, gao2023selfguided} focus on the use of BERT or GPT-2 for generating data, but their model design and fine-tuning are often challenging due to the limitations of data diversity and the need to not completely detach from the distribution of the original dataset\citep{pouran-ben-veyseh-etal-2023-generating}. However, methods based on the GPT3.5 family no longer require fine-tuning and it is more close to an end-to-end approach to data generation. Especially in low-resource domains \citep{xu-etal-2023-s2ynre}, GPT3.5 is trained on massive corpora and has a vast data pool, and this a priori knowledge greatly contributes to the accuracy of the generated data in text classification \citep{dai2023auggpt, piedboeuf-langlais-2023-chatgpt, tang2023does} and RE \citep{yoo-etal-2021-gpt3mix-leveraging, xu2023unleash}.

\section{Method}
\subsection{Problem Definition}
The goal of a RE task is to extract the relation between several pairs of entities mentioned in a sentence based on a given sentence. Typically a RE dataset consists of a corpus of sentences, entity pairs, and relation types. An RE sample contains a sentence $S=\left\{w_{l} \right\}^{L}_{l=1} $ consisting of multiple words $w $, and pairs of head entities $e_{s} $ and tail entities $e_{o} $ mentioned by $S $, where $e_{s}, e_{o} \in \mathbb{E} $ and relation $r \in \mathbb{R} $, where $\mathbb{E} $ and $\mathbb{R} $ represent a set of predefined entity types and relation types, respectively. 

Specifically, the RE task means that given a sentence $S $, the model needs to predict, from $\mathbb{R} $, the correspondence (if it exists) between $e_{s}$ and $e_{o}$ contained in $S $, obtain the relation triplet $\left(e_{s}, r, e_{o}\right)$. That is, the model needs to learn each training sample $\mathcal{D}=\left\{ \left( S, e_{s_{i}}, e_{o_{i}} \right) \right\}^{N}_{i=1}$ to predict its corresponding relation label $r_{i}$:
\begin{displaymath}
\label{eq1}
	p_{LM}\left( r|S, e_{s}, e_{o}\right)=\prod\limits_{i=1}^N p\left(r_{i}|S, e_{s_{i}}, e_{o_{i}} \right)
\end{displaymath}
The model predicts the probability distribution of the set of possible relations and determine the relation with the highest probability based on each training sample $\mathcal{D} $:
\begin{displaymath}
\label{eq2}
r^{\ast}=\mathop{\arg\max}\limits_{r \in \mathbb{R}} \prod\limits_{i=1}^N p\left( r_{i}|S, e_{s_{i}}, e_{o_{i} } \right)
\end{displaymath}

\subsection{Span-level Entity and Relation Extraction}
We aims to address the span-based ERE task, all entity information given in the RE task is predicted by the NER module in the model.Specifically, span-based means formally that a given sentence $S $ consists of multiple tokens: $S=\left\{t_{i} \right\}^{n}_{i=1} $. A span is defined as a continuous sequence of tokens, and spans have start and end indices representing the boundaries, define $Span= \left\{ x_{1}, x_{2}, x_{3}, ... , x_{m} \right\} $ as all possible spans, entities are defined as spans labeled with entity types, and relations are defined as pairs of entity spans labeled with relation types. For the NER task, the model needs to predict the entity type of each span, map each span to an entity type: $y_{e}(x_{i}) \in \mathbb{E} $ represents that the span is an entity with entity type, while $y_{e}(x_{i}) \notin \mathbb{E} $ represents that the span is not an entity, the output of the NER is $Y_{e}=\{(x_{i}, e):x_{i}\in Span,e\in \mathbb{E}\}$. For the RE task, the model needs to predict the relation type for each pair of entity spans$x_{s}, x_{o} \in Span $, each pair of entity spans is mapped to a relation type: $y_{r}(x_{s}, x_{o}) \in \mathbb{R} $ represents that the pair of entity spans has relation type, $y_{r}(x_{s}, x_{o}) \notin \mathbb{R} $ represents that they have no relation, the output of the RE is $Y_{r}=\{(x_{s}, x_{o},r):s_{s},s_{o}\in Span,r\in\mathbb{R}\}$.

\subsection{Method Overview}
\begin{figure*}[t]
  \includegraphics[width=16cm]{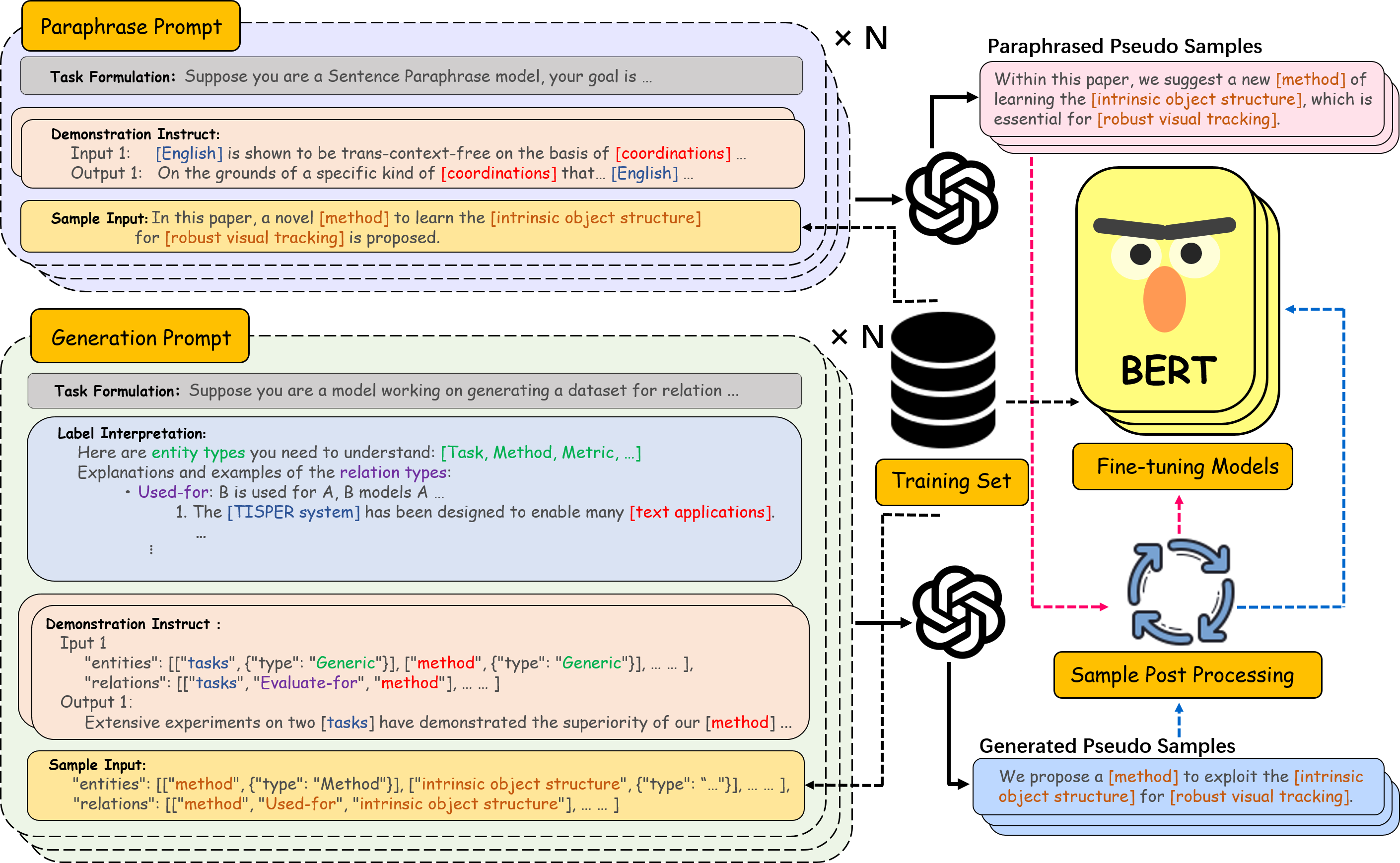}
  \centering
  \caption{General framework of the PGA. The left part of the Figure shows the prompt composition of the framework for both two data augmentation approaches. The two prompt containing the original training samples are iteratively input into GPT-3.5, respectively, to synthesize two kind of pseudo-samples, which are post-processed, filtered, and converted into the format required by each ERE model to engage in fine-tuning along with the original training samples.}
  \label{fig:overview}
\end{figure*}

As shown in Fig~\ref{fig:overview}, we propose a data augmentation framework PGA (Paraphrasing-based and Generating-based Augmentation) for RE in the scientific domain, which guides LLM to synthesize pseudo-samples through paraphrasing and generating approaches to enhance the performance of RE models. Specifically, the data augmentation prompt is designed into easy-to-understand sample data format. Through these well-designed prompt, LLM synthesize pseudo-samples with label information. Then, after data post-processing step to filter the wrong samples to control the sample quality and transform the pseudo-samples into the original data format, these additional pseudo-samples are utilized together with the original training dataset to fine-tune the respective backbone RE models.

\subsection{Paraphrasing-based Data Augmentation Approach}
For each sample $\mathcal{D}=\left\{ \left( S, e_{s_{i}}, e_{o_{i}} \right) \right\}^{N}_{i=1}$ of the original training set, where the relation semantics is implied in the sentence text $S $, the boundary of entity span$\mathcal{E^{\prime}}= \left\{ e_{i}^{\prime} \right\}^{N}_{i=1}$ (without entity type labels) of $S $ is inputted to the LLM. Through sentence paraphrasing, LLM is instructed to synthesize a new text $S_{P} $,  which implies the same relation semantics, but is different from the original sentence representation and form. Finally, concatenate the original entity information to obtain a pseudo-sample $\mathcal{D_{P}}=\left\{ \left( S_{P}, e_{s_{i}}, e_{o_{i}} \right) \right\}^{N}_{i=1} $. The pseudo-sample $\mathcal{D_{P}} $ allows the RE model to learn more information, including more expressions and presentations of the same relation type and increases the probability of correctly predicting the relation type. By considering the process of text paraphrasing as a text sequence generation task for string $y $, the process can be written as follows:
\begin{displaymath}
p_{LLM}\left( S_{P}|S, e_{s}^{\prime}, e_{o}^{\prime} \right)=\prod\limits_{l=1}^L p\left(y_{l}|S, e_{s}^{\prime}, e_{o}^{\prime}, y_{<l} \right)
\end{displaymath}
That is the probability of using LLM to generate a string $y $ (paraphrased sentence) of length $L $.

\subsection{Generating-based Data Augmentation Approach}
The entity information $\mathcal{E}= \left\{ e_{i}\right\}^{N}_{i=1} $ corresponding to sentence $S $ in each sample $\mathcal{D}=\left\{ \left( S, e_{s_{i}}, e_{o_{i}} \right) \right\}^{N}_{i=1} $ of the original training set, together with the final result to be predicted by the model, i.e., the relation label$\mathcal{R}=\left\{ r_{i} \right\}^{N}_{i=1} $, are fed into the LLM. With the powerful text sequence generation capability of LLM, the sentence $S_{G} $ containing all the entities as well as relational semantics are obtained. As mentioned above, the purpose of RE is to predict the type of relation $\mathcal{R} $ implicit in the sample sentences by learning each training sample $\mathcal{D} $ by the model. And generating-based data augmentation approach can be viewed as a partial reverse-order process for this task - i.e., the LLM generates learnable samples based on the provided answer labels. For each sample, go through all its known labeling information $\mathcal{M}=\left\{e_{s_{i}}, e_{o_{i}}, r_{i}\right\}^{N}_{i=1}$ to generate a new sentence $S_{G} $, and finally concatenate it with the original entity information to get a pseudo-sample $\mathcal{D}_{G}=\left\{ \left( S_{G}, e_{s_{i}}, e_{o_{i}} \right) \right\}^{N}_{i=1} $. These generated sentences in samples present different descriptions and representations than the original sentences, implying full labeling information, and making the model to learn more additional in-domain labeling samples. Similarly, by considering the generating approach as a text sequence generation task for string $y $, the process can be written as follows:
\begin{displaymath}
p_{LLM}\left( S_{G}|e_{s}, e_{o}, r \right)=\prod\limits_{l=1}^L p\left(y_{l}|e_{s}, e_{o}, r, y_{<l} \right)
\end{displaymath}

\subsection{Prompt Construction}
As shown in the left part of Fig~\ref{fig:overview}, the framework constructs prompt for each given training set sample for the two data augmentation approaches, and the complete generated prompt are detailed in the Appendix \ref{sec:appendix}.
\\
\textbf{Task Formulation} \quad Instruct LLM to perform role-playing of a RE sample paraphrasing or generating model and stipulate the task definition and objectives. Detail precautions to prevent LLM from synthesizing non-compliant samples, and adding or deleting entity information, so that LLM can further clarify the requirements of the task. For generating approach, we elaborate the format of the input text. Specify the genre and style of the output text to make it fit more closely to the original sentences. Since the entity information input may be empty, we specifies that the pseudo-samples without label information are "No result can be generated with the given information.", and its entity and relation information are empty.
\\
\textbf{Demonstration Instruct} \quad For LLM to better learn the specified data format and understand the task, also limited by the length of the prompt, we provide two random sample as demonstration. 
In practice, we find LLM's understanding of sentences is affected by the presence of too much structured information in the input data samples. Because most of the priori knowledge it learned during the pre-training phase is text corpus, that form is easy to be read by humans, so its way of thinking is closer to the human. Since the backbone models are span-based, which lead to the sentence in tokens and the label data in JSON format, nested with lots of brackets, quotation marks. Too much complex and useless information will make the sample information lost. Therefore, rather than using the dataset's JSON format, we make a simple adaptation of the input and output sample formats. The tokens are reverted to a complete sentence, and the entity in the sentence is wrapped in square brackets without entity type $\mathbb{E} $. We merely utilizes the boundary information of the entity span ($\mathcal{E^{\prime}}= \left\{ e_{i}^{\prime} \right\}^{N}_{i=1}$). This shortens the repetitive information and facilitates LLM comprehension and learning. For generating approaches, the outputs use the adapted format. However, the input adopts JSON format, the input are not sentences and entity information, but complete entity and relation labels.
\\
\textbf{Label Interpretation} \quad Generating-based Data augmentation needs to ensure that the samples still maintain the labeling accuracy of entities and relations, keeping the quality of the samples as close as possible to the original data. However, it is also important to avoid generating overly simple and repetitive text to ensure that the model can learn new information and patterns. The prompt provide all entity label sets, relation type label and interpretation, and several cases of each relation type to constrain the LLM to avoid that the meaning of the text it generates does not match with the corresponding label, which affects the quality of the pseudo-sample. Among them, the description of relation labels is derived from \citet{luan-etal-2018-multi}.
\\
\textbf{Sample Input} \quad This prompt contains information of a sample from the original training dataset. Each data sample needs to be iteratively inserted into the prompt with reference to the format in the example, and then repeatedly input into the LLM.

\subsection{Data Post-processing}
After obtaining the sentences output from the LLM, in the data post-processing step, the sentences are merged with their entity and relation label information of the original dataset to form a complete pseudo-sample. Also, it compares the information with the original samples, identifies and filters out erroneous pseudo-samples, and ensures that the data is accurate. Finally, the filtered pseudo-samples are reverted to the format that can be read by each RE model.

\section{Experiment}
\subsection{Dataset}
we evaluate the PGA framework on a scientific domain RE dataset. The SciERC \citep{luan-etal-2018-multi} dataset collects 500 abstracts of AI papers, labeling them with relation and entity information. We synthesized a large number of paraphrased and generated pseudo-samples, the statistics of the datasets it synthesized are shown in Table~\ref{tab:dataset}.
\begin{table*}[h]
\begin{center}
\resizebox{\textwidth}{!}{
\begin{tabular}{ccccccc}
   \toprule
   \bf Dataset & \bf \#Train & \bf \#Dev & \bf \#Test & \bf \#Ents(Types) & \bf \#Rels(Types) & \bf \#Defect Rate\\
   \midrule
   $SciERC$ & 1,861 & 275 & 551 & 8,089(6) & 4,716(7) & - \\
   $SciERC_{P}$ & 1,861 & - & - & 5,598(6) & 3,219(7) & 21.60\% \\
   $SciERC_{G}$ & 1,589 & - & - & 4,341(6) & 2,402(7) & 14.61\% \\
   \bottomrule
\end{tabular}
}
\end{center}
\caption{\label{font-table} Statistics of original dataset, paraphrased ($SciERC_{P}$) and generated ($SciERC_{G}$) datasets}
\label{tab:dataset}
\end{table*}

\subsection{Evaluation Metrics}
For the NER task, entity span boundaries, and types need to be correctly predicted. For the RE task, we reports two evaluation metrics (1) \textbf{Boundary evaluation (Rel)} requires the model to correctly predict the span boundaries of each subject and object entity pair and the relation type between them; (2) \textbf{Strict evaluation (Rel+)} in the setting of Rel further requires the model to correctly predict the types of entities as well \citep{taille-etal-2020-lets}.

\subsection{Baseline and Backbone Model}
To evaluate the PGA framework, the following state-of-the-art RE models are selected as the backbone of the PGA framework: (1) \textbf{SpERT} \citep{Eberts2019SpanbasedJE} is the mainstream joint ERE model, which enumerates all possible fragment spans in a text, encodes and generates vector representations of spans, and performs entity labeling classification prediction of the spans. All extracted entities span are paired and the relation types between entity pairs are determined by a multi-classification module. (2) \textbf{PL-Marker} \citep{ye-etal-2022-packed} is a pipeline model and proposes a neighborhood-oriented packing strategy for the NER module to model entity boundary information more accurately. A subject-object entity packing strategy is designed for RE, which uses solid and levitated markers so that different object entities can share the same subject entity. (3) \textbf{PURE} \citep{zhong-chen-2021-frustratingly} is an end-to-end pipeline model, where the NER and RE modules are trained independently and share the same pre-trained encoder. They use markers for the RE and handles each pair of subject and object entities independently. In addition, we use six mainstream RE models as baseline for comparison.

\subsection{Implementation Details}
For LLM, OpenAI's state-of-the-art GPT-3.5-based long text processing model is \emph{text-davinci-003}. The practice shows that \emph{gpt-3.5-turbo} is more applicable to the chat conversation domain than to the long text generation domain, we choose \emph{text-davinci-003} as the the base LLM of the PGA framework.

For paraphrasing-based data augmentation, GPT-3.5 temperature is set to 0.5, while for the generating-based approach, due to the need for a better diversity of samples, the temperature is set to 1. All the samples from the original training set are paraphrased and generated, though the GPT-3.5 outputs incorrect samples in practice. Since the paraphrased sentences possess the original sentences as a clear answer to contrast, the incorrect paraphrased samples are filtered and screened out in the data post-processing step, and then paraphrased samples with errors are re-synthesized until the correct sample is obtained. While for generating approach, the input is label information, which is hard to control, so the incorrect sample is directly discarded. The defect rates of the pseudo-samples synthesized by the two approaches are detailed in Table~\ref{tab:dataset}. When reducing the pseudo-sentences to the token format, we use the program to uniformly generate pseudo-attribute to replace irrelevant attribute information (e.g., the source of the sentences in the sample, i.e., ID) that are required by the models but are not involved in the NER and RE tasks.

For backbone models, for SpERT \citep{Eberts2019SpanbasedJE}, \emph{scibert-scivocab-cased}\citep{beltagy-etal-2019-scibert} is used as the encoder, and the model is trained on SciERC's both training and development sets according the original work; For PL- Marker \citep{ye-etal-2022-packed} and PURE \citep{zhong-chen-2021-frustratingly}, \emph{scibert-scivocab-uncased} \citep{beltagy-etal-2019-scibert} is used as the encoder and the models are trained on SciERC's training set, as per the original work. The hyperparameters are optimized for the backbone models in the main and combining pseudo-sample experiments, while the other experiments are conducted according to the hyperparameters of the original work.

\subsection{Main Results}
The results of the PGA framework were compared with the performance of each backbone model and baseline methods, and the results are shown in Table\ref{tab:main}. Among the two data augmentation methods of PGA, $PGA_{P} $ denotes paraphrasing-based data augmentation and $PGA_{G} $ denotes generating-based data augmentation. From the table, it can be learned that data augmentation is a simple and effective method for RE, which can borrow the prior knowledge of LLM to synthesize pseudo-samples to train fine-tuning-based models to enhance the performance. The PGA framework achieves the highest F1 scores under the three metrics of Ent, Rel, and Rel+, which also outperforms all the previous baselines. In terms of backbone model comparison, $PGA_{P} $ based on SpERT \citep{Eberts2019SpanbasedJE} achieves the best F1 scores in Ent and Rel+ with the highest improvement of 1.78\%, while $PGA_{G} $ achieves 1.53\% F1 score improvement in Rel+ compared to the original, but with some decline in Rel metrics; $PGA_{G} $ based on PL- Marker \citep{ye-etal-2022-packed} compared to the original, achieves improvement in every metric, with the highest F1 score improvement of 2.05\% in Rel+, and $PGA_{G} $ likewise improves in all metrics, achieving the best of all models in Ent and Rel metrics, with 1.26\% improvement in F1 scores in both Rel and Rel+; $PGA_{P} $ based on PURE \citep{zhong-chen-2021-frustratingly} also all metrics improves, Rel improves by 1.81\%, Rel+ improves by 1.67\%, and also $PGA_{G} $ improves in all metrics, Rel+ improves by up to 1.9\%.

Meanwhile, we notice that the performance of very few metrics even decrease. Generally, the PGA framework has improvements in RE task, especially under strict metrics.

\begin{table*}[h]
\begin{center}
\resizebox{\textwidth}{!}{
\begin{tabular}{ccccccccccc}
   \toprule
   \multirow{2}{*}{\bf Model}  & \multirow{2}{*}{\bf Encoder}  & \multicolumn{3}{c}{\bf Ent} & \multicolumn{3}{c}{\bf Rel} & \multicolumn{3}{c}{\bf Rel+} \\ \cmidrule(lr){3-5}\cmidrule(lr){6-8}\cmidrule(lr){9-11}
   \multicolumn{2}{c}{~} & Precision & Recall & \bf F1 & Precision & Recall & \bf F1 & Precision & Recall & \bf F1\\
   \midrule
   SciIE \citep{luan-etal-2018-multi} & BiLSTM & 67.20 & 61.50 & 64.20 & 47.60 & 33.50 & 39.30 & - & - & - \\
   DyGIE \citep{luan-etal-2019-general} & ELMo & - & - & 65.20 & - & - & 41.60 & - & - & - \\
   DyGIE++ \citep{wadden-etal-2019-entity} & $BERT_{Base} $ & - & - & 67.50 & - & - & 48.40 & - & - & - \\
   UniRE(\citep{wang-etal-2021-unire}) & \multirow{12}{*}{SciBERT} & 65.8 & 71.10 & 68.40 & - & - & $40.20^{\dagger} $ & 37.30 & 36.60 & 36.90 \\
   TriMF \citep{shen2021triggersense} & \multirow{12}{*}{~} & 70.18 & 70.17 & 70.17 & 52.63 & 52.32 & 52.44 & - & - & - \\
   PFN \citep{yan-etal-2021-partition} & \multirow{12}{*}{~} & - & - & 66.80 & - & - & - & - & - & 38.40 \\ \cmidrule(lr){1-1}\cmidrule(lr){3-11}
   SpERT$^{\ast} $ \citep{Eberts2019SpanbasedJE} & \multirow{12}{*}{~} & 68.54 & 70.09 & \underline{69.31} & 52.66 & 50.82 & \bf 51.72 & 39.15 & 37.78 & 38.45 \\
   SpERT+$PGA_{P} $ & \multirow{12}{*}{~} & 69.43 & 69.67 & \textbf{69.55} & 50.72 & 50.51 & \underline{50.62} & 40.31 & 40.14 & \bf{40.23} \\
   SpERT+$PGA_{G} $ & \multirow{12}{*}{~} & 68.72 & 69.38 & 69.05 & 52.27 & 47.33 & 49.68 & 42.06 & 38.09 & \underline{39.98} \\ \cmidrule(lr){1-1}\cmidrule(lr){3-11}
   PL-Marker \citep{ye-etal-2022-packed} & \multirow{12}{*}{~}  & - & - & 69.90 & - & - & 53.20 & - & - & 41.60 \\
   PL-Marker+$PGA_{P} $ & \multirow{12}{*}{~} & 71.15 & 68.78 & \underline{69.95} & - & - & \underline{53.40} & - & - & \bf 43.65 \\
   PL-Marker+$PGA_{G} $ & \multirow{12}{*}{~} & 70.11 & 70.86 & \bf 70.48 & - & - & \bf 54.46 & - & - & \underline{42.86} \\ \cmidrule(lr){1-1}\cmidrule(lr){3-11}
   PURE$^{\ast} $ \citep{zhong-chen-2021-frustratingly} & \multirow{12}{*}{~} & 69.33 & 68.55 & 68.93 & 49.69 & 49.69 & 49.69 & 37.37 & 37.37 & 37.37 \\
   PURE+$PGA_{P} $ & \multirow{12}{*}{~} & 70.46 & 68.07 & \bf{69.24} & 55.09 & 48.36 & \bf{51.50} & 41.75 & 36.65 & \underline{39.04} \\
   PURE+$PGA_{G} $ & \multirow{12}{*}{~} & 70.50 & 67.95 & \underline{69.21} & 55.58 & 44.97 & \underline{49.72} & 43.91 & 35.52 & \bf{39.27} \\
   
   \bottomrule
\end{tabular}
}
\end{center}
\caption{\label{font-table} 
Results of the PGA framework plus each backbone model on the SciERC test set. The performance is evaluated in terms of accuracy, recall, and F1 score. \textbf{Bold} indicates the highest performance under each backbone model, and data with \underline{underline} indicates sub-optimal performance. Please note: Models with $\ast $ indicate that to obtain as many evaluation indicators as possible, we rerun the experiment according the setting and code published by the original work. Result with $\dagger $ indicate replicated results obtained by \citet{ren-etal-2023-covariance} on the original model.}
\label{tab:main}
\end{table*}

\section{Analysis}
\subsection{Combine two Types of Pseudo-samples}
Combine two types of Pseudo-samples, alongside the original training set to fine-tune backbone models. The results of the combining pseudo-samples $PGA_{C} $ are shown in Table~\ref{tab:combine}. On SpERT \citep{Eberts2019SpanbasedJE}, $PGA_{C} $ performs poorly, and most performance metrics show different degrees of decline; on PURE \citep{zhong-chen-2021-frustratingly}, even all performance metrics are declined; only on PL-Marker \citep{ye-etal-2022-packed} Rel+ has 0.99\% improvement. We speculate whether this is due to the annotation quality of the pseudo-samples is problematic. As can be seen from Table~\ref{tab:dataset}, the size of the combining dataset far exceeds the original training dataset, which may introduce a large number of unlabeled entities. And the lack of understanding of the vocabulary of the scientific dataset generates much noise that the quality of the entire dataset is lower than the original training set, which leads to a degradation of the performance of the noise-sensitive model.

\begin{table*}[h]
\begin{center}
\resizebox{\textwidth}{!}{
\begin{tabular}{cccccccccc}
   \toprule
   \multirow{2}{*}{\bf Model} & \multicolumn{3}{c}{\bf Ent} & \multicolumn{3}{c}{\bf Rel} & \multicolumn{3}{c}{\bf Rel+} \\ \cmidrule(lr){2-4}\cmidrule(lr){5-7}\cmidrule(lr){8-10}
   ~ & Precision & Recall & \bf F1 & Precision & Recall & \bf F1 & Precision & Recall & \bf F1\\
   \midrule
    SpERT$^{\ast} $ \citep{Eberts2019SpanbasedJE} & 68.54 & 70.09 & \bf 69.31 & 52.66 & 50.82 & \bf 51.72 & 39.15 & 37.78 & 38.45 \\
    SpERT+$PGA_{C} $ & 69.40 & 67.72 & 68.55 & 53.00 & 44.46 & 48.35 & 42.72 & 35.83 & \bf 38.97 \\ \hline
    PL-Marker \citep{ye-etal-2022-packed} & - & - & 69.90 & - & - & \bf 53.20 & - & - & 41.60 \\
    PL-Marker+$PGA_{C} $ & 70.61 & 69.44 & \bf 70.02 & - & - & 52.83 & - & - & \bf 42.59 \\ \hline
    PURE$^{\ast} $ \citep{zhong-chen-2021-frustratingly} & 69.33 & 68.55 & \bf 68.93 & 49.69 & 49.69 & \bf 49.69 & 37.37 & 37.37 & \bf 37.37 \\
    PURE+$PGA_{C} $ & 69.61 & 67.41 & 68.50 & 49.62 & 47.43 & 48.50 & 37.16 & 35.52 & 36.33 \\
   \bottomrule
\end{tabular}
}
\end{center}
\caption{\label{font-table} The performance of combining two types pseudo-samples with the original dataset for fine-tuning the backbone models, \textbf{bold} indicates the highest performance under each backbone model. $PGA_{C} $ denotes that each backbone model is trained by combining datasets.}
\label{tab:combine}
\end{table*}

\begin{figure*}[t]
  \includegraphics[width=16cm]{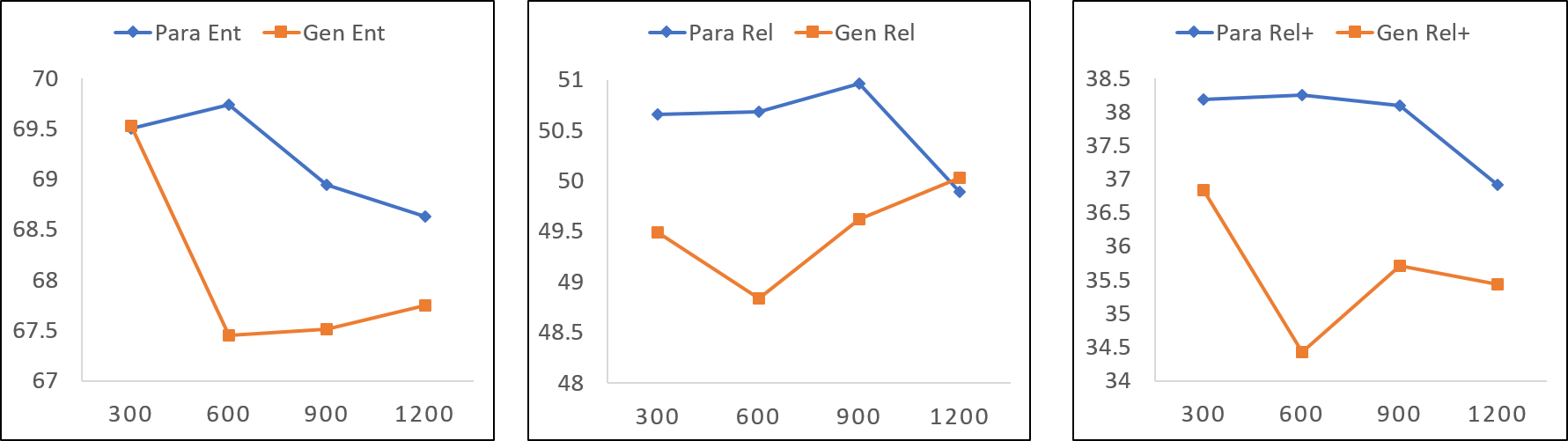}
  \centering
  \caption{Performance variation of SpERT \citep{Eberts2019SpanbasedJE} model with different numbers of pseudo samples and original training set involved in training, the horizontal coordinate represents the number of pseudo samples involved and the vertical coordinate represents the F1 score.}
  \label{fig:number}
\end{figure*}

\subsection{Quality of Pseudo-samples}
Remove the original dataset and use only the pseudo-samples alone to fine-tune each backbone model, and the results are shown in Table~\ref{tab:sole}. Overall, the performance of each model that is fine-tuned using paraphrased and generated pseudo-samples alone shows a decrease. These disparities indicate that the quality of the pseudo-samples has some room for improvement. Even in the best-case scenario, SpERT \citep{Eberts2019SpanbasedJE} trained with the paraphrased pseudo-samples, shows a decrease of 3.06\% in Ent F1 socre, a decrease of 5.88\% in Rel F1 score, and a decrease of 4.12\% in Rel+. However, on PURE \citep{zhong-chen-2021-frustratingly}, the F1 scores decrease the most among all models in generated pseudo-samples training, with Ent's F1 score decreasing by 25.56\%, Rel's F1 score even decreasing by 29.34\%, and the Rel+ score dropping by 22.95\%. The decline of PL-Marker \citep{ye-etal-2022-packed} is somewhere in between.

Overall, the quality of the paraphrased pseudo-samples is higher than the generated in the framework. Every F1 score of the backbone model with the training of the paraphrased pseudo-samples exceeds the generated, and each metric differs by more than 10\%. We consider the main reasons is that for the entity, the entity information relies on the input samples to provide, while LLM introduce a certain amount of noisy entities. However, for relation, the paraphrasing approach synthesizes relation information almost relying on LLM's inference of relations from understanding of the meaning of the sentence, and it is challenging for LLM to deduce the relation implied in the text. In addition, LLM synthesizes relation information under more stringent conditions with generating-based method, with only types of label information and no sentences for reference. This is very less constraining, so the LLM does not introduce more implicit relation semantics in the sentences, and the samples retain a certain amount of noise, resulting in lesser quality than its predecessor.

\begin{table*}[h]
\begin{center}
\resizebox{\textwidth}{!}{
\begin{tabular}{cccccccccc}
   \toprule
   \multirow{2}{*}{\bf Model} & \multicolumn{3}{c}{\bf Ent} & \multicolumn{3}{c}{\bf Rel} & \multicolumn{3}{c}{\bf Rel+} \\ \cmidrule(lr){2-4}\cmidrule(lr){5-7}\cmidrule(lr){8-10}
   ~ & Precision & Recall & \bf F1 & Precision & Recall & \bf F1 & Precision & Recall & \bf F1\\
   \midrule
    SpERT$^{\ast} $ \citep{Eberts2019SpanbasedJE} & 68.54 & 70.09 & 69.31 & 52.66 & 50.82 & 51.72 & 39.15 & 37.78 & 38.45 \\
    SpERT+$PGA^{Sole}_{P} $ & 67.24 & 65.28 & 66.25 & 48.67 & 43.33 & 45.84 & 36.45 & 32.44 & 34.33 \\
    SpERT+$PGA^{Sole}_{G} $ & 50.34 & 56.74 & 53.35 & 36.73 & 21.46 & 27.09 & 26.01 & 15.20 & 19.18 \\ \hline
    PL-Marker \citep{ye-etal-2022-packed} & - & - & 69.90 & - & - & 53.20 & - & - & 41.60 \\
    PL-Marker+$PGA^{Sole}_{P} $ & 67.00 & 59.41 & 62.98 & - & - & 44.68 & - & - & 34.63 \\
    PL-Marker+$PGA^{Sole}_{G} $ & 52.67 & 41.61 & 46.49 & - & - & 26.60 & - & - & 19.42 \\ \hline
    PURE$^{\ast} $ \citep{zhong-chen-2021-frustratingly} & 69.33 & 68.55 & 68.93 & 49.69 & 49.69 & 49.69 & 37.37 & 37.37 & 37.37 \\
    PURE+$PGA^{Sole}_{P} $ & 59.25 & 55.13 & 57.12 & 40.37 & 31.42 & 35.33 & 29.95 & 23.31 & 26.21 \\
    PURE+$PGA^{Sole}_{G} $ & 52.21 & 37.09 & 43.37 & 34.62 & 14.68 & 20.62 & 24.21 & 10.27 & 14.42 \\
   \bottomrule
\end{tabular}
}
\end{center}
\caption{\label{font-table} 
Performance of SpERT \citep{Eberts2019SpanbasedJE} with only pseudo-samples involved in training. $PGA^{Sole}_{P} $ denotes only paraphrased pseudo-samples engage in training and $PGA^{Sole}_{G}$ denotes only generated pseudo-samples engage in training.}
\label{tab:sole}
\end{table*}

\subsection{Performance of Fine-tuning with Different Numbers of Pseudo-samples}
Splitting the two approaches' pseudo-samples separately, incorporating each of them into the original training set in varying quantities to determine if the number of pseudo-samples also affect on the performance, the results are shown in Figure~\ref{fig:number}. For NER, as the number of paraphrased pseudo-samples rises, the performance initially improves briefly and then decreases again due to the addition of too many pseudo-samples. Generated pseudo-samples, however, undergoes an initial dip and then rises slowly.

For RE, the performance for both Rel and Rel+ peak at a specific number as the number of paraphrased pseudo-samples is boosted, and then gradually decreases as more pseudo-samples are added, at which point the size of the pseudo-samples is close to the original training set. The overall performance trend of RE is not dissociated from the change in performance of NER. This is also reflected in the RE performance with generated pseudo-samples. Overall, Rel+ performance again declines with more pseudo-samples involved, so we infer that the performance of NER and RE does not consistently improve with the addition of more pseudo-samples.

\begin{figure*}[t]
  \includegraphics[width=16cm]{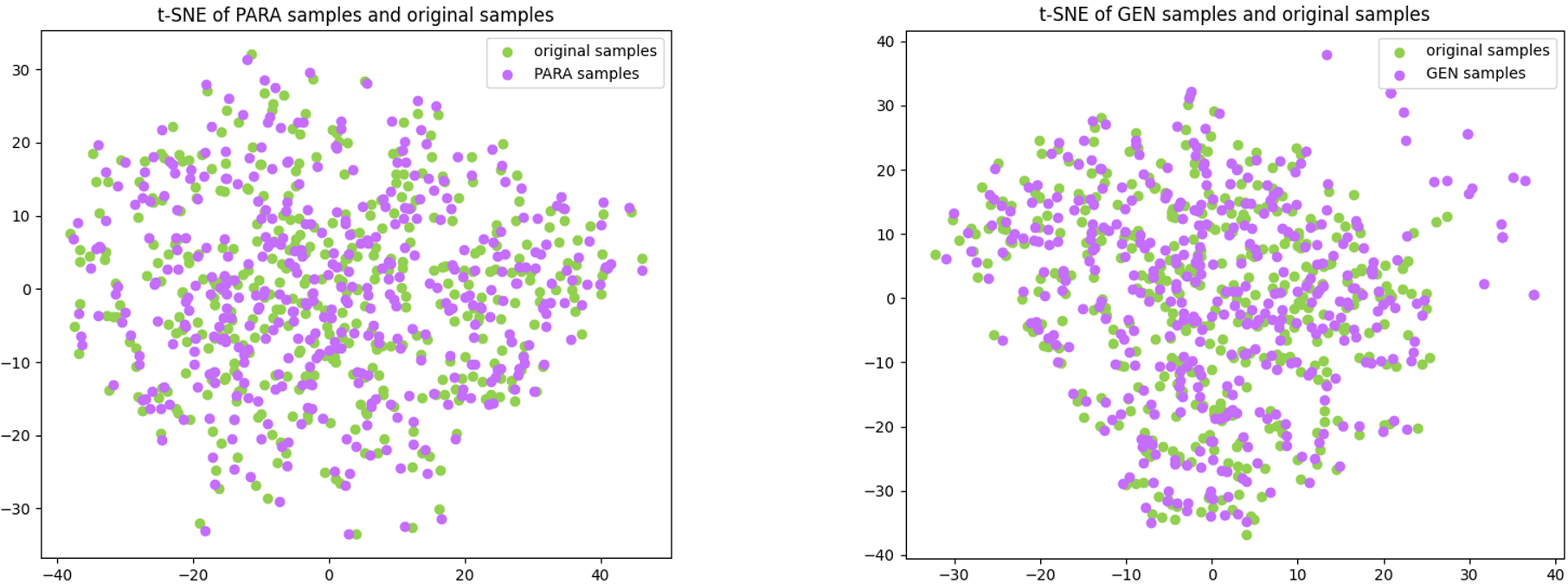}
  \centering
  \caption{Distribution of the embeddings of the pseudo-samples and the sentences of the original dataset in the vector space.}
  \label{fig:fidelity}
\end{figure*}

\subsection{Fidelity of Pseudo-samples}
To measure whether the pseudo-samples are close to the real original training set samples, we use the \emph{all-MiniLM-L6-v2} model in \emph{sentence-transformers} to textually embed 400 paraphrased and generated pseudo-sentences, together with the corresponding original sentences, respectively. Then use t-SNE algorithm to downscale and visualize the high-dimensional sentence embedding in the vector space to obtain the spatial distribution of the embedding. As can be seen in Figure~\ref{fig:fidelity}, the paraphrased pseudo-samples maintain a relatively high degree of overlap with the original samples. For every original sample, even the sporadic sample at the edges of the figure, there is a very close paraphrased pseudo-sample. However, the generated pseudo-samples roughly overlap and they are weaker than the former in terms of closeness. Furthermore, from the upper right corner of its distribution, it can be observed that the pseudo sample distribution is far away from the corresponding original sample. Therefore, the generated samples are less faithful to the original samples than the paraphrased samples.

\section{Conclusion}
In this work, we propose a data augmentation framework based on a GPT-3.5 called PGA, which can synthesize pseudo-samples according to the original dataset through two data augmentation methods, namely, paraphrasing and generating method. On a dataset of more challenging scientific domains, we perform data augmentation on the mainstream RE model to improve the performance of the model. After conducting analytical experiments on the pseudo-samples, we concludes that the pseudo-samples paraphrased by the PGA are effective in introducing more semantic information of the same relation for model learning while maintaining a higher fidelity to the original samples. 

\section*{Acknowledgments}
Thanks to the Open Fund Project of the State Key Laboratory of Intelligent Manufacturing System Technology "Research on complex product performance inference and evaluation method based on multi-source data mining" for supporting this work.

\bibliography{custom}

\appendix
\section{Appendix}
\label{sec:appendix}
\newpage

\begin{table*}[h]
\begin{center}
\begin{tabular}{p{3cm} p{12cm}}
   \toprule
   \multicolumn{2}{c}{\bf Paraphrase Prompt}\\
   \midrule
   
   \bf Task Formulation & Suppose you are a sentence paraphrase model. Your goal is to follow a given sentence and the entities in the sentence, describe the sentence in different words than the original sentence while keeping the entities in the sentence and the meaning of the sentence intact, then output it.
    ~
   
   \textbf{Note:} In the sentence to be processed, the part enclosed in parentheses is the entity; do not add, delete, or change the entity, or output any redundant information. Please fully understand the meaning of the sentence before paraphrasing.

    ~
   
    \\
    \bf Demonstration Instruct & Please refer to the following examples for the format of the output, and please study the following examples to accomplish the above tasks more accurately.

    ~
    
    \textbf{Iput1:} '[English] is shown to be trans-context-free on the basis of [coordinations] of the respectively type that involve [strictly syntactic cross-serial agreement].'
    
    \textbf{Output1:} On the grounds of a specific kind of [coordinations] that require [strictly syntactic cross-serial agreement], [English] is demonstrated to be trans-context-free.

    ~

    \textbf{Iput2:} 'The [agreement] in question involves number in [nouns] and [reflexive pronouns] and is syntactic rather than semantic in nature because [grammatical number] in [English], like [grammatical gender] in [languages] such as [French] , is partly arbitrary.'
    
    \textbf{Output2:} The number in [nouns] and [reflexive pronouns] that must match in this [agreement] is based on syntax rather than meaning, since [English] [grammatical number], like [grammatical gender] in [languages] like [French], is partly random.

    ~

    \\
    \bf Sample Input & \textbf{Sentence:} In this paper we show how two standard [outputs] from [information extraction (IE) systems]-[named entity annotations] and [scenario templates]-can be used to enhance access to [text collections] via a standard [text browser].\\

   \bottomrule
\end{tabular}
\end{center}
\caption{The detailed paraphrase prompt}
\end{table*}

\begin{table*}[h]
\begin{center}
\begin{tabular}{p{3cm} p{12cm}}
   \toprule
   \multicolumn{2}{c}{\bf Paraphrase Prompt}\\
   \midrule
   
   \bf Task Formulation & Suppose you are a model working on generating a dataset for relation extraction. One data sample of a relation extraction dataset is a sentence, and its corresponding entity information and relation information. Entities are embedded in the sentence, entities definitely have entity types, and certain pairs of entities will also contain corresponding relation types based on semantics. \textbf{Your goal:} to generate sentences that contain information about entities and relations, given that information.
   
    ~
   
    \\

   \bf Label Interpretation & Here is the information you need to understand: 6 entity types: \textbf{[Task, Method, Metric, Material, Generic, OtherScientificTerm]}
   
   Below are explanations and examples of the relation types, with pairs of entities where a relation exists enclosed in parentheses.
   
    \textbf{• Used-for:} B is used for A, B models A, A is trained on B, B exploits A, A is based on B.E.g. 
    
    1. The [TISPER system] has been designed to enable many [text applications]. 
    
    2. Our [method] models [user proficiency]. 
    
    3. Our [algorithms] exploits [local soothness]. 

    \textbf{• Feature-of:} B belongs to A, B is a feature of A, B is under A domain. E.g. 
    
    1. [prior knowledge] of the [model]
    
    2. [genre-specific regularities] of [discourse structure]
    
    3. [English text] in [science domain] 

    \textbf{• Hyponym-of:} B is a hyponym of A, B is a type of A.E.g.
    
    1. [TUIT] is a [software library]
    
    2. [NLP applications] such as [machine translation] and [language generation]

    \textbf{• Part-of:} B is a part of A... E.g. 
    
    1. The [system] includes two models: [speech recognition] and [natural language understanding]
    
    2. We incorporate [NLU module] to the [system]. 
    
    \textbf{• Evaluate-for:} B is evaluated for A, A use B to get estimation, E.g.
    
    1. [Intra-sentential quality] is evaluated with [rule-based heuristics]
    
    2. We describe a new [system] that enhances Criterion's capability , by evaluating multiple aspects of [coherence in essays].

    \textbf{• Compare:} Opposite of conjunction, compare two models/methods, or listing two opposing entities. E.g.
    
    1. Unlike the [quantitative prior], the [qualitative prior] is often ignored...
    
    2. We compare our [system] with previous [sequential tagging systems]...

    \textbf{• Conjunction:} Function as similar role or use/incorporate with. E.g.
    
    1. obtained from [human expert] or [knowledge base]
    
    2. NLP applications such as [machine translation] and [language generation]\\

   \bottomrule
\end{tabular}
\end{center}
\end{table*}

\begin{table*}[h]
\begin{center}
\begin{tabular}{p{3cm} p{12cm}}
   \toprule
   \multicolumn{2}{c}{\bf Generate Prompt (Continued)}\\
   \midrule
    \bf Demonstration Instruct &  \textbf{Caution: }Do not leave out any entities, if the 'entities' is empty, output 'No result can be generated with the given information.', and please try to avoid generating entities in the output sentence that are not provided by the input, such as 'Problem', 'Task', 'Method' and so on. Do not leave out any relation or add relation between pairs of entities that don't have given relation. The genre of each sample sentence generated is the abstract of an AI paper. Please study the following examples and refer to them for the output format (entities enclosed in square brackets in the output sentences)

    ~

    \textbf{Input1:}\{\textbf{'entities':} [['agreement', \{'type': 'Generic'\}], ['nouns',\{'type': 'OtherScientificTerm'\}], ['reflexive pronouns', \{'type': 'OtherScientificTerm'\}], ['grammatical number', \{'type': 'OtherScientificTerm'\}], ['English', \{'type': 'Material'\}], ['grammatical gender', \{'type': 'OtherScientificTerm'\}], ['languages', \{'type': 'Material'\}], ['French', \{'type': 'Material'\}]], \textbf{'relations':} [['nouns', 'Conjunction', 'reflexive pronouns'], ['grammatical gender', 'Feature-of', 'languages'], ['French', 'Hyponym-of', 'languages']]\}
    
    \textbf{Output1: }'The [agreement] in question involves number in [nouns] and [reflexive pronouns] and is syntactic rather than semantic in nature because [grammatical number] in [English], like [grammatical gender] in [languages] such as [French] , is partly arbitrary.'

    ~
    
    \textbf{Input2:}\{\textbf{'entities':} [['tasks', \{'type': 'Generic'\}], ['method', \{'type': 'Generic'\}], ['state-of-the-art methods', \{'type': 'Generic'\}]], 
    \textbf{'relations':} [['tasks', 'Evaluate-for', 'method'], ['method', 'Compare', 'state-of-the-art methods']]\}
    
    \textbf{Output2:}'Extensive experiments on two [tasks] have demonstrated the superiority of our [method] over the [state-of-the-art methods].'

    ~
    
    \\
    \bf Sample Input & Now please generate the output based on my input.
    
    \textbf{Input: }\{\textbf{'entities':} [['method', \{'type': 'Method'\}], ['intrinsic object structure', \{'type': 'OtherScientificTerm'\}], ['robust visual tracking', \{'type': 'Task'\}]], \textbf{'relations':} [['method', 'Used-for', 'intrinsic object structure'], ['intrinsic object structure', 'Used-for', 'robust visual tracking']]\}
    \\

   \bottomrule
\end{tabular}
\end{center}
\caption{The detailed generate prompt}
\end{table*}

\end{document}